# ENHANCEMENT OF IMAGES USING MORPHOLOGICAL TRANSFORMATIONS

## K.Sreedhar[1] and B.Panlal[2]

[1]Department of Electronics and communication Engineering, VITS (N9)
Karimnagar, Andhra Pradesh, India
`sreedhar_kallem@yahoo.com`

[2]Department of Electronics and communication Engineering, VCE (S4)
Karimnagar, Andhra Pradesh, India
`panlal4u@yahoo.com`

## ABSTRACT

*This paper deals with enhancement of images with poor contrast and detection of background. Proposes a frame work which is used to detect the background in images characterized by poor contrast. Image enhancement has been carried out by the two methods based on the Weber's law notion. The first method employs information from image background analysis by blocks, while the second transformation method utilizes the opening operation, closing operation, which is employed to define the multi-background gray scale images. The complete image processing is done using MATLAB simulation model.*

*Finally, this paper is organized as follows as Morphological transformation and Weber's law. Image background approximation to the background by means of block analysis in conjunction with transformations that enhance images with poor lighting. The multibackground notion is introduced by means of the opening by reconstruction shows a comparison among several techniques to improve contrast in images. Finally, conclusions are presented.*

## 1. INTRODUCTION

The image enhancement problem in digital images can be approached from various methodologies, among which is mathematical morphology (MM). Such operators consist in accordance to some proximity criterion, in selecting for each point of the analyzed image, a new grey level between two patterns (primitives) [1], [4]. Even though morphological contrast has been largely studied, there are no methodologies, from the point of view MM, capable of simultaneously normalizing and enhancing the contrast in images with poor lighting. On the other side, one of the most common techniques in image processing to enhance dark regions is the use of nonlinear functions, such as logarithm or power functions ; otherwise, a method that works in the frequency domain is the homomorphism filter. However, the main disadvantage of histogram equalization is that the global properties of the image cannot be properly applied in a local context, frequently producing a poor performance in detail preservation. In a method to enhance contrast is proposed; the methodology consists in solving an optimization problem that maximizes the average local contrast of an image.

This paper deals with the detection of background in images with poor contrast. The complete image processing is done using MATLAB simulation model.

The optimization formulation includes a perceptual constraint derived directly from human super threshold contrast sensitivity function. The authors apply the proposed operators to some images with poor lighting with good results. On the other hand a methodology to enhance contrast based on colour statistics from a training set of images which look visually appealing is presented. Here, the basic idea is to select a set of training images which look good perceptually, next a Gaussian mixture model for the colour distribution in the face region is built, and for any given input image, a colour tone mapping is performed so that the colour statistics in the face region matches the training examples[2],[3],[5]. In this way, even though the reported algorithms to compensate changes in lighting are varied, some are more adequate than others. In this work, two methodologies to compute the image background are proposed. Also, some operators to enhance and normalize the contrast in grey level images with poor lighting are introduced. Contrast operators are based on the logarithm function in a similar way to Weber's law the use of the logarithm function avoids abrupt changes in lighting. Also, two approximations to compute the background in the processed images are proposed. The first proposal consists in an analysis by blocks, whereas in the second proposal, the opening by reconstruction is used given its following properties:

- It passes through regional minima, and
- It merges components of the image without considerably modifying other structures.

Finally, this paper is organized as follows [8]. Morphological transformation and Weber's law presents a brief background on Weber's law and some morphological transformations. Image background approximation to the background by means of block analysis in conjunction with transformations that enhance images with poor lighting. The multibackground notion is introduced by means of the opening by reconstruction. A comparison among several techniques to improve contrast in images. Finally, conclusions are presented. The aim of the paper is to detect the background image and enhance the contrast in gray level image with poor lighting. First operator applies information from block analysis and second operator's uses opening by reconstruction.

## 1.1. Existing System

The range of intensity i.e. the difference between highest and lowest intensity values in an image gives a measure of its contrast. There are standard techniques like histogram equalization, histogram stretching for improving the poor contrast of the degraded image [11]. However there is a need for devising context-sensitive techniques based on local contrast variation since the image characteristics differ considerably from one region to another in the same image and also the local histogram does not necessarily follow the global histogram. The enhancement level is not significant and provides good results only for certain images but fails to provide good results for most of the images, especially those taken under poor lighting. In other words, it doesn't provide good performance for detail preservation.

## 1.2. Proposed System

In a method to enhance contrast is proposed; the methodology consists in solving an optimization problem that maximizes the average local contrast of an image. The optimization formulation includes a perceptual constraint derived directly from human threshold contrast sensitivity function. The authors apply the proposed operators to some images with poor lighting with good results. On the other hand a methodology to enhance contrast based on color statistics from a training set of images which look visually appealing is presented. Here, the basic idea is to select a set of training images which look good perceptually, next a Gaussian mixture model for the color distribution in the face region is built, and for any given input image, a color tone mapping is performed so that the color statistics in the face region matches the training examples. In this way, even though the reported algorithms to compensate changes in lighting are varied, some are more adequate than others.

# 2. MORPHOLOGICAL TRANSFORMATIONS AND WEBER'S LAW

## 2.1. Morphology

Morphology is a technique of image processing based on shape and form of objects. Morphological methods apply a structuring element to an input image, creating an output image at the same size. The value of each pixel in the input image is based on a comparison of the corresponding pixel in the input image with its neighbors. By choosing the size and shape of the neighbor, you can construct a morphological operation that is sensitive to specific shapes in the input image. The morphological operations can first be defined on grayscale images where the source image is planar (single-channel). The definition can then be expanded to full-colour images.

## 2.2. Morphological Operations

Morphological operations such as erosion, dilation, opening, and closing. Often combinations of these operations are used to perform morphological image analysis [3], [17]. There are many useful operators defined in mathematical morphology. They are dilation, erosion, opening and closing. Morphological operations apply structuring elements to an input image, creating an output image of the same size. Irrespective of the size of the structuring element, the origin is located at its centre. Morphological opening is $\gamma_{\mu B}(f)(x)$ and Morphological closing is $\varphi_{\mu B}(f)(x)$

$$\gamma_{\mu B}(f)(x) = \delta_{\mu \tilde{B}}(\varepsilon_{\mu B}(f))(x)$$
$$\varphi_{\mu B}(f)(x) = \varepsilon_{\mu \tilde{B}}(\delta_{\mu B}(f))(x) \tag{1}$$

where $\mu$ a homothetic parameter, size is $\mu$ means a square of $(2\mu+1) \times (2\mu+1)$ pixels. B is the structuring element of size $3 \times 3$ (here $\mu = 1$).

### 2.2.1. Dilation

Dilation is a transformation that produces an image that is the same shape as the original, but is a different size. Dilation stretches or shrinks the original figure [10]. Dilation increases the valleys and enlarges the width of maximum regions, so it can remove negative impulsive noises but do little on positives ones.

The dilation of A by the structuring element B is defined by:

$$A \oplus B = \underset{b \in B}{U} A_b \tag{2}$$

If B has a center on the origin, as before, then the dilation of A by B can be understood as the locus of the points covered by B when the center of B moves inside A.

Dilation of image f by structuring element s is given by $f \oplus s$. The structuring element s is positioned with its origin at *(x, y)* and the new pixel value is determined using the rule:

$$g(x, y) = \begin{cases} 1 & if \ s \ hits \ f \\ 0 & otherwise \end{cases} \tag{3}$$

The following figure illustrates the morphological dilation of a gray scale image. Note how the structuring element defines the neighbourhood of the pixel of interest, which is circled. The dilation function applies the appropriate rule to the pixels in the neighbourhood and assigns a value to the corresponding pixel in the output image. In the figure, the morphological dilation function sets the value of the output pixel to 16 because it is the maximum value of all the pixels in the input pixel's neighbourhood defined by the structuring element is on. Easiest way to describe it is to imagine the same fax/text is written with a thicker pen

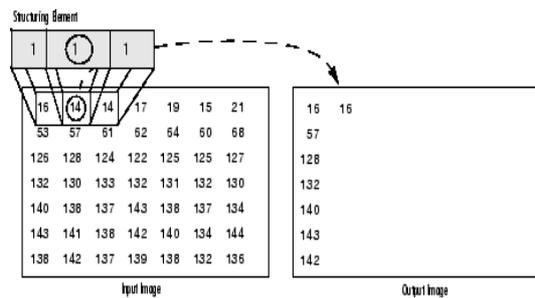

Figure 1.Morphological Dilation of a Gray scale Image

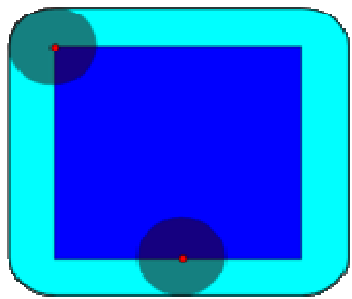

Figure 2. Example for Dilation operation

In a binary image, if any of the pixels is set to the value 1, the output pixel is set to 1.

### 2.2.2.  Erosion

It is used to reduce objects in the image and known that erosion reduces the peaks and enlarges the widths of minimum regions, so it can remove positive noises but affect negative impulsive noises little.

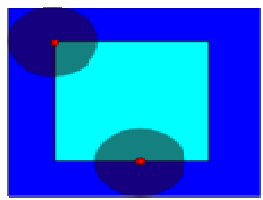

Figure 3. Example for Erosion operation

The erosion of the dark blue square resulting in light blue square.
The erosion of the binary image $A$ by the structuring element $B$ is defined by:

$$A \ominus B = \left\{ z \in E \middle| B_z \subseteq A \right\} \tag{4}$$

In a binary image, if any of the pixels is set to 0, the output pixel is set to 0.

Erosion of image *f* by structuring element *s* is given by    $f \ominus s$. The structuring element s is positioned with its origin at *(x, y)* and the new pixel value is determined using the rule:

$$g(x, y) = \begin{cases} 1 & if \; s \;\; hits \;\; f \\ 0 & otherwise \end{cases} \qquad (5)$$

In the above equation fit means all *on pixel* in the structuring element covers an *on pixel* in the image.

The following figure illustrates the morphological erosion of a gray scale image. Note how the structuring element defines the neighbourhood of the pixel of interest, which is circled. The dilation function applies the appropriate rule to the pixels in the neighbourhood and assigns a value to the corresponding pixel in the output image. In the figure, the morphological erosion function sets the value of the output pixel to 14 because it is the minimum value of all the pixels in the input pixel's neighbourhood defined by the structuring element is on.

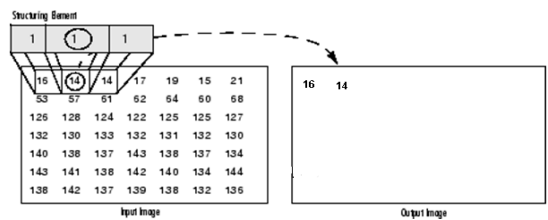

Figure 4. Morphological Erosion of a Grayscale Image

### 2.2.3. Opening Operation

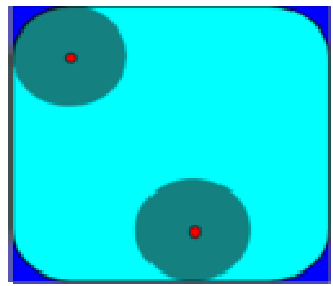

Figure 5. Opening

The opening of *A* by *B* is obtained by the erosion of *A* by *B*, followed by dilation of the resulting image by *B*:

$$A \circ B = (A \; ! \;\; B) \oplus B \qquad (6)$$

In the case of the square of side 10, and a disc of radius 2 as the structuring element, the opening is a square of side 10 with rounded corners, where the corner radius is 2.

The sharp edges start to disappear. Opening of an image is erosion followed by dilation with the same structuring element.

### 2.2.4. Closing Operation

Closing of an image is the reverse of opening operation.

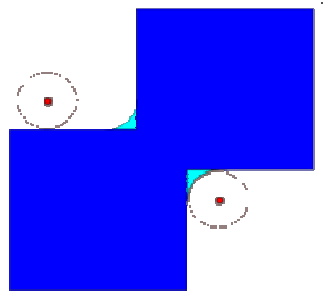

Figure 6.Closing

The closing of *A* by *B* is obtained by the dilation of *A* by *B*, followed by erosion of the resulting structure by *B*:

$$A \bullet B = (A \oplus B) \ominus B \qquad (7)$$

The first method proposed is the block analysis where the entire image is split into a number of blocks and each block is enhanced individually. The next proposed method is the erosion-dilation method which is similar to block analysis but uses morphological operations (erosion and dilation) for the entire image rather than splitting into blocks. All these methods were initially applied for the gray level images and later were extended to colour images by splitting the colour image into its respective R, G and B components, individually enhancing them and concatenating them to yield the enhanced image. All the above mentioned techniques operate on the image in the spatial domain. The final method is the DCT where the frequency domain is used. Here we scale the DC coefficients of the image after DCT has been taken. The DC coefficient is adjusted as it contains the maximum information. Here, we move from RGB domain to YCbCr domain for processing and in YCbCr, to adjust (scale) the DC coefficient, i.e. Y (0, 0). The image is converted from RGB to YCbCr domain because if the image is enhanced without converting, there is a good chance that it may yield an undesired output image. The enhancement of images is done using the log operator [1].This is taken because it avoids abrupt changes in lighting. For example, if 2 adjacent pixel values are 10 and 100, their difference in normal scale is 90.But in the logarithmic scale, this difference reduces to just 1, thus providing a perfect platform for image enhancement.

## 2.3. Weber's Law

The study of contrast sensitivity has dominated visual perception research. In psycho-visual studies, the contrast C of an object with luminance $L_{max}$ against its surrounding luminance $L_{min}$ is defined as follows [13], [14], and [15]:

$$C = \frac{L_{max} - L_{min}}{L_{min}} \qquad (8)$$

$C$ – Contrast of the image, $L_{max}$ —Luminance of the image and $L_{min}$ – Luminance of the surroundings

If $L = L_{min}$ and $\Delta L = L_{max} - L_{min}$ , (8) can be rewritten as [21]

$$C = \frac{\Delta L}{L} \qquad (9)$$

On the other hand, in, a methodology to compute the background parameter was proposed. The methodology consists in calculating the average between the smallest and largest regional minima. However, the main disadvantage of this proposal is that the image background is not detected in a local way. As a result, the contrast is not correctly enhanced in images with poor lighting, since considerable changes occur in the image background due to abrupt changes in luminance.

In this paper, an approximation to Weber's law [14] is considered by taking the luminance L as the grey level intensity of a function (image); Equation (9) indicates that $\Delta(\log L)$ is proportional to $C$; therefore Weber's law can be expressed as

$$C = k \log L + b \qquad L > 0 \tag{10}$$

This law has a logarithmic relation. This technique is applied to image processing to enhance the image effectively. Where 'C' is the contrast, 'k' and 'b' are constants, 'b' being the background parameter and 'k' being the scaling factor for enhancement. Weber's law can be best understood from the following example. Consider a photo taken in a dark room. The obtained photo actually consists of 2 different things. One is what we visually perceive in that image and the other is what is actually present in that image. Weber's law simply states that the relation between these two is logarithmic.

In our case, an approximation to Weber's law is considered by taking the luminance L as the grey level intensity of a function (image); in this way, expression (10) is written as

$$C = k \log f + b \qquad f > 0 \tag{11}$$

# 3. IMAGE BACKGROUND ANALYSIS BY BLOCKS

Morphological transformations (Opening by reconstruction, Erosion-Dilation method) and Block Analysis is used to detect the background of gray level and colour images [3]. These techniques are first implemented in gray scale and are then extended to colour images by individually enhancing the colour components. For aiding better results, the compressed domain (DCT) technique is used exclusively for colour image enhancement. The major advantage of the DCT method is that it can be used for any type of illumination. In all the above methods, the enhancement of the background detected image is done using Weber's law (modified Weber's law for compressed domain) a critical analysis of the various advantages and drawbacks in each method are performed and ways for overcoming the drawbacks are also suggested. Here, the results of each technique are illustrated for various backgrounds, majority of the minimum poor lighting condition [6]. In image acquisition, background detection is necessary in many applications to get clear and useful information from an image which may have been picturized in different conditions like poor lighting or bright lighting, moving or still etc. This Section deals with background analysis of the image by blocks. In this project, $D$ represent the digital space under study, with $D = Z * Z$ and $Z$ and $Z$ is the integer set. For each analyzed block, maximum $(M_i)$ and minimum $(m_i)$ values are used to determine the background measures. $\tau_i$ is used to select the background parameters. Background parameters lie between clear ($f > \tau_i$) and dark ($f \leq \tau_i$) intensity levels. If ($f \leq \tau_i$) is the dark region then background parameters takes the maximum intensity levels $(M_i)$ then ($f > \tau_i$) is the clear region, background parameters takes the minimum intensity levels $(m_i)$.

Enhance images are we get after applying the below equation [18],

$$\Gamma_{\tau(x)}(f) = \begin{cases} k_{\tau(x)} \log(f+1) + \delta_\mu(f)(x), & f \leq \tau(x) \\ k_{\tau(x)} \log(f+1) + \varepsilon_\mu(f)(x), & Otherwise \end{cases} \tag{12}$$

and

$$k_{\tau(x)} = \frac{255 - \tau(x)}{\log(256)} \tag{13}$$

$\delta$ - Dilation operation, $\varepsilon$ - Erosion operation

Dilation and erosion are the two most common morphological operations used for back ground analysis by blocks.

### 3.1. Block Analysis for Gray level images

Let $f$ be the original image which is subdivided into number of blocks with each block is the sub-image of the original image. For each and every block n, the minimum intensity $m_i$ and maximum intensity $M_i$ values are calculated $m_i$ and $M_i$ values are used to find the background criteria $\tau_i$ in the following way [20]:

$$\tau_i = \frac{m_i + M_i}{2} \qquad \forall i = 1, 2......n \tag{14}$$

$\tau_i$ is used as a threshold between clear ($f > \tau_i$) and dark ($f = \tau_i$) intensity levels. Based on the value of $\tau_i$, the background parameter is decided for each analyzed block. Correspondingly the contrast enhancement is expressed as follows:

$$\Gamma_{\tau_i}(f) = \begin{cases} k_i \log(f+1) + M_i, & f \le \tau_i \\ k_i \log(f+1) + m_i, & Otherwise \end{cases} \tag{15}$$

It is clear that the background parameter entirely is dependent up on the background criteria $\tau_i$ value. For $f = \tau_i$, the background parameter takes the maximum intensity value $M_i$ within the analyzed block, and the minimum intensity value $m_i$ otherwise. In order to avoid in determination condition, unit was added to the logarithmic function [12], [19].

$$Where \, k_i = \frac{255 - m_i^*}{\log(256)} \qquad \forall i = 1, 2,.....,n \tag{16}$$

$$With \, m_i^* = \begin{cases} m_i, & f \le \tau_i \\ M_i, & f \ge \tau_i \end{cases} \tag{17}$$

The more is the number of blocks; the better will be quality of the enhanced image. In the enhanced images, it can be seen that the objects that are not clearly visible in the original image are revealed. As the size of the structuring element increases it is hard to preserve the image as blurring and contouring effects are severe. The results are best obtained by keeping the size of the structuring element as 2 (μ=2). Sample input (left half of the image) and output image (right half) for block analysis is shown below:

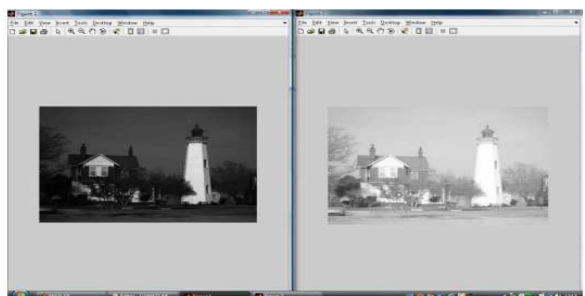

Figure 7. Block Analysis (μ=2)

# 4. IMAGE BACKGROUND ANALYSIS USING OPENING BY RECONSTRUCTION FOR GRAY SCALE IMAGES

This method is similar to block analysis in many ways; apart from the fact that the manipulation is done on the image as a whole rather than partitioning it into blocks. Firstly minimum $I_{min}(x)$ and maximum intensity $I_{max}(x)$ contained in a structuring element (B) of elemental size $3 \times 3$ is calculated.

The above obtained values are used to find the background criteria $\tau_i$ as described below [19]

$$\tau(x) = \frac{I_{min}(x) + I_{max}(x)}{2} \qquad (18)$$

Where $I_{min}(x)$ and $I_{max}(x)$ corresponds to morphological erosion and dilation respectively, Therefore

$$\tau(x) = \frac{\varepsilon_\mu(f)(x) + \delta_\mu(f)(x)}{2} \qquad (19)$$

In this way the contrast operator can be described as in equations (12) and (13).

By employing Erosion-Dilation method we obtain a better local analysis of the image for detecting the background criteria than the previously used method of Blocks. This is because the structuring element μB permits the analysis of eight boring pixels at each point in the image. By increasing the size of the structuring element more pixels will be taken into account for finding the background criteria [9]. It can be easily visualized that several characteristics that are not visible at first sight appear in the enhanced images. The trouble with this method is that morphological erosion or dilation when used with large size of μ to reveal the background, undesired values maybe generated.

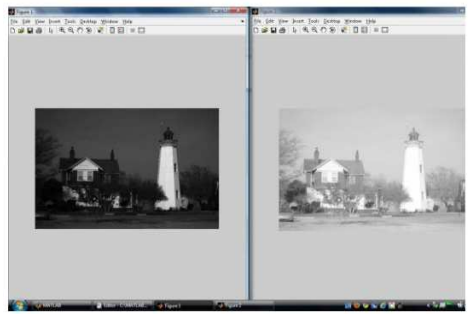

Figure 8. Closing By Reconstruction (μ=2)

In general it is desirable to filter an image without generating any new components. The transformation function which enables to eliminate unnecessary parts without affecting other regions of the image is defined in mathematical morphology which is termed as transformation by reconstruction. We go for opening by reconstruction because it restores the original shape of the objects in the image that remain after erosion as it touches the regional minima and merges the regional maxima. This particular characteristic allows the modification of the altitude of regional maxima when the size of the structuring element increases thereby aiding in detection of the background criteria as follows:

$$\tau(x) = \tilde{\gamma}_{\mu B}(f)(x) \qquad (20)$$

Where opening by reconstruction is expressed as

$$\bar{\gamma}_{\mu B}(f)(x) = \lim_{n \to \infty} \delta_f^n(\varepsilon_{\mu B}(f))(x) \tag{21}$$

It can be observed from the above equation that opening by reconstruction first erodes the input image and uses it as a marker. Here marker image is defined because this is the image which contains the starting or seed locations. For example, here the eroded image can be used as the marker. Then dilation of the eroded image i.e. marker is performed iteratively until stability is achieved.

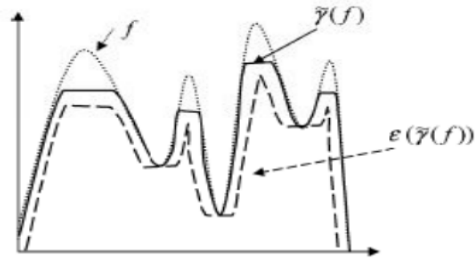

Figure 9. Image background obtained from the erosion of the opening by reconstruction

Background parameter b(x) is calculated by eroding the above obtained background criterion τ(x) which is described below:

$$b(x) = \varepsilon_1[\bar{\gamma}_\mu(f)](x) \tag{22}$$

As it is already mentioned that morphological erosion [22], [23] will generate unnecessary information when the size of the structuring element is increased, in this study, the image background was calculated by choosing the size of the structuring element as unity.

Contrast enhancement is obtained by applying Weber's law as expressed below [2], [7]:

$$\xi_{\bar{\gamma}\mu}(f) = k(x)\log(f+1) + \varepsilon_1[\bar{\gamma}_\mu(f)] \tag{23}$$

and

$$k(x) = \frac{\max \mathrm{int} - \varepsilon_1[\bar{\gamma}_\mu(f)]}{\log(\max \mathrm{int} + 1)} \tag{24}$$

Where, $\max \mathrm{int}$ refers to maximum gray level intensity which is equal to 255. If the intensity of the background increases, the image becomes lighter because of the additive effect of the whiteness (i.e. maximum intensity) of the background. It is to be remembered that it is the objective of opening by reconstruction to preserve the shape of the image components that remain after erosion.

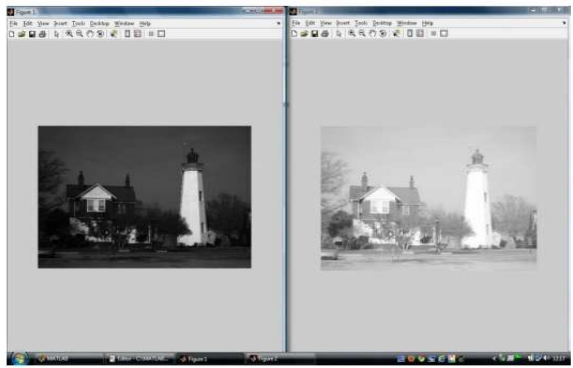

Figure10. Opening by Reconstruction (μ=2)

# 5. SIMULATION RESULTS

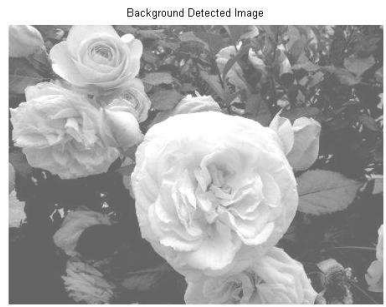

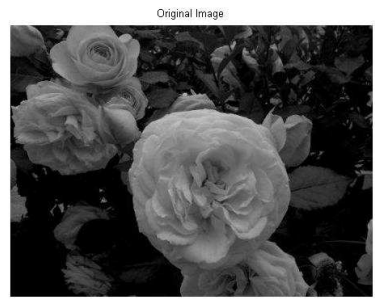

(a1)                                            (a2)

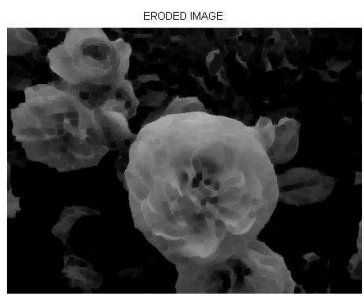

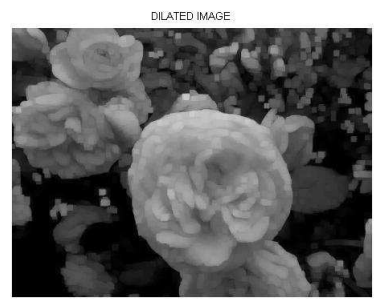

(a3)                                            (a4)

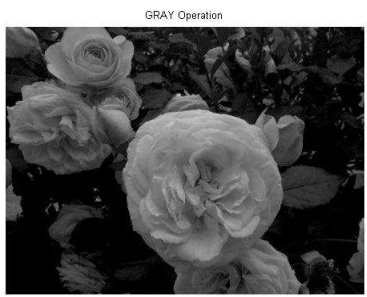

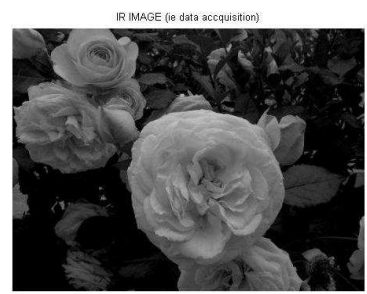

(a5)                                            (a6)

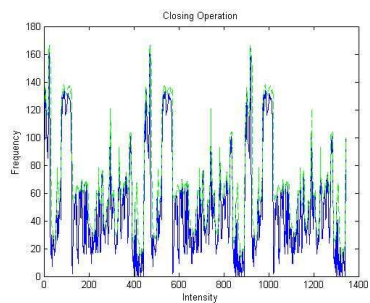

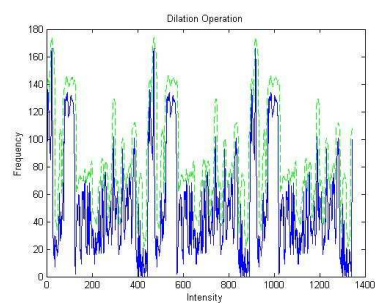

(a7)                                            (a8)

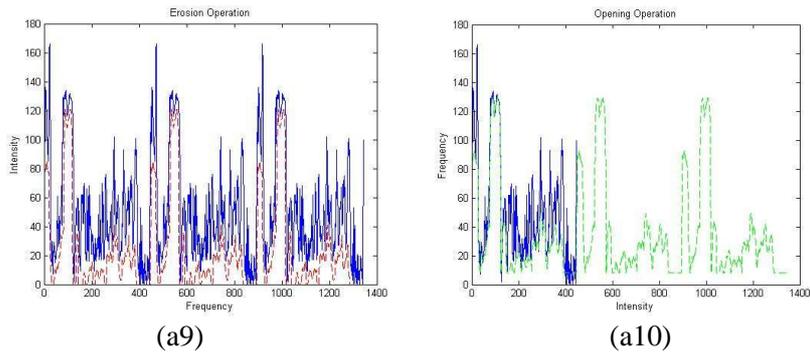

(a9)               (a10)

Figure 11. Image background detection using the morphological erosion and dilation. (a1).Background detection image, (a2). Original image (a3).Eroded image, (a4). Dilated image (a5).GRAY operation image, (a6). IR image (a7).Closing Operation, (a8). Dilation Operation (a9).Erosion Operation, (a10). Opening Operation

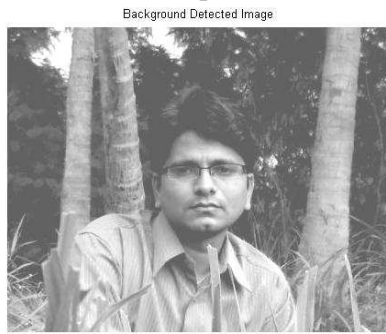

(b1)

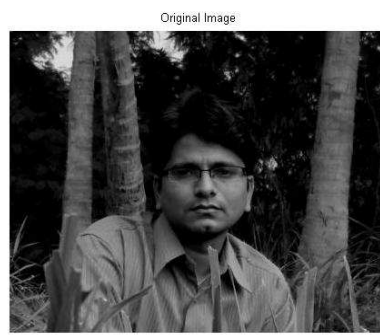

(b2)

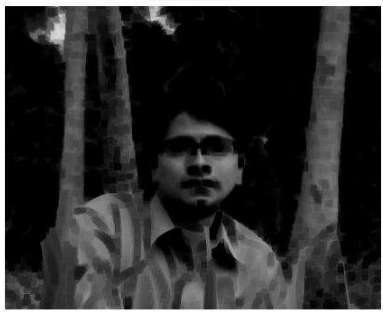

(b3)

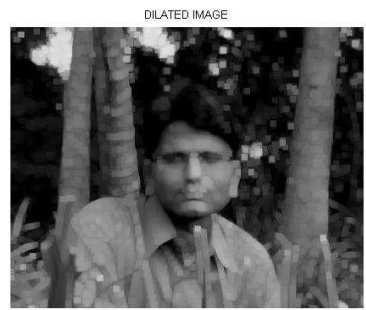

(b4)

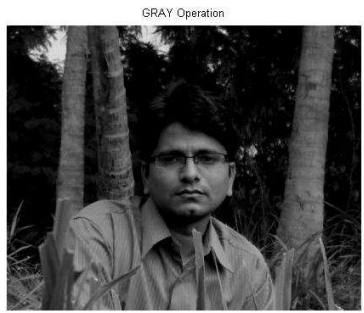

(b5)

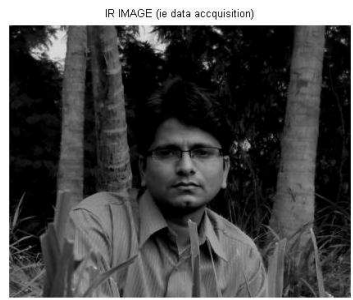

(b6)

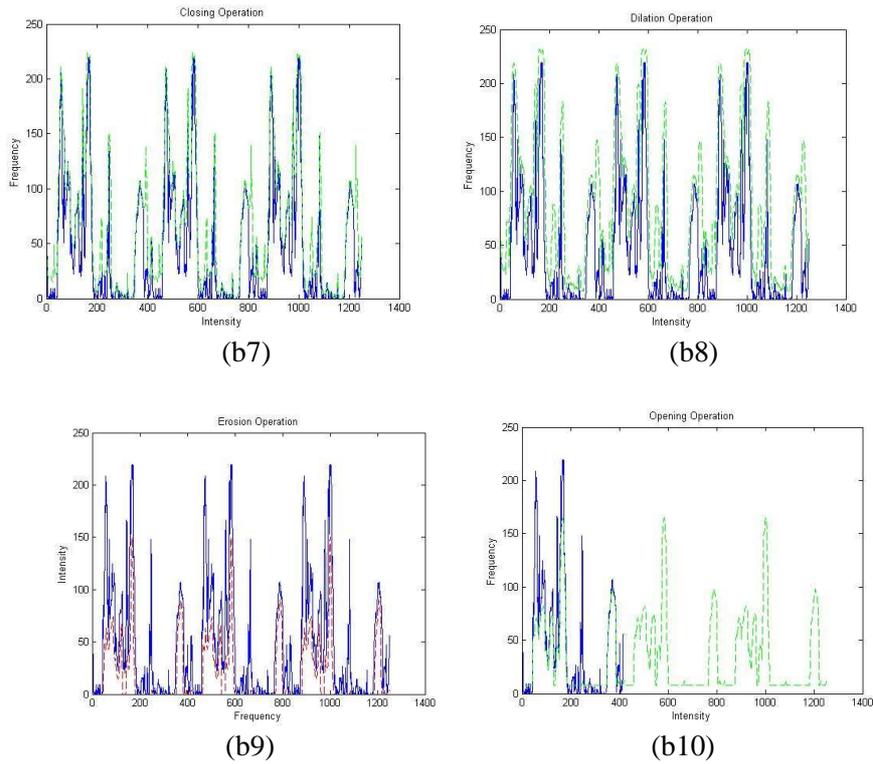

Figure 12.Image background obtained from the opening by reconstruction
(b1).Background detection image, (b2). Original image (b3).Eroded image, (b4). Dilated image
(b5).GRAY operation image, (b6). IR image     (b7).Closing Operation, (b8). Dilation Operation
(b9).Erosion Operation, (b10). Opening Operation

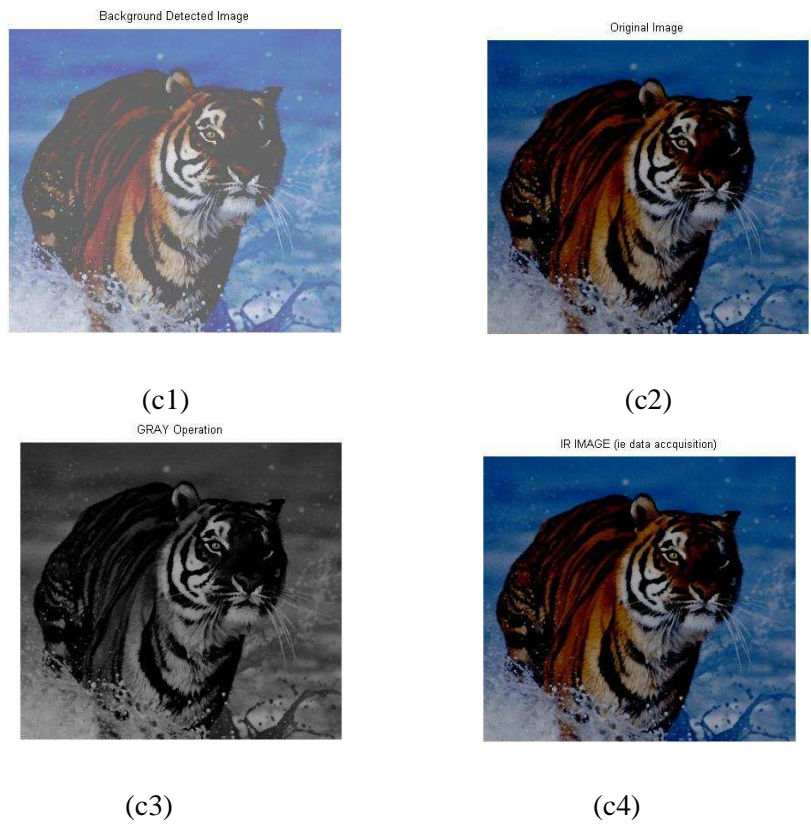

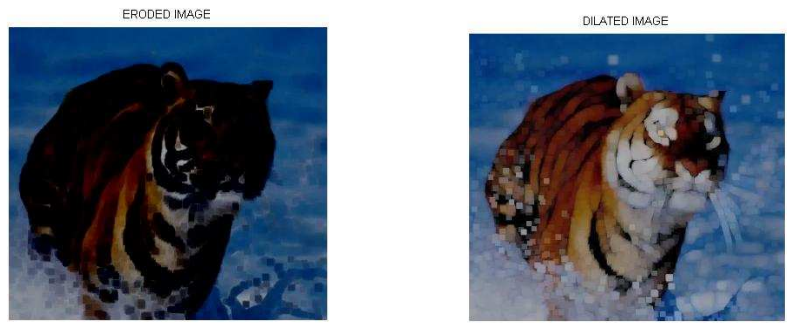

(c5)                                                    (c6)

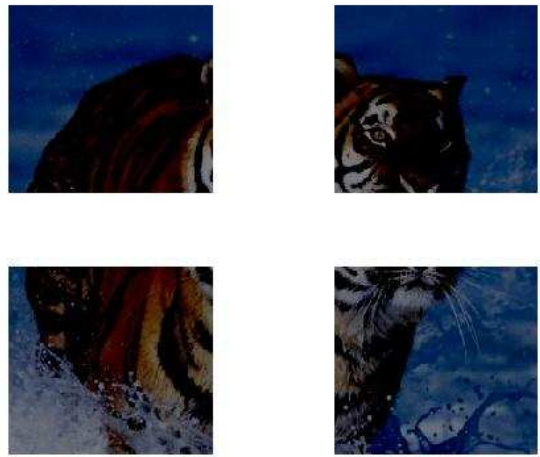

(c7)

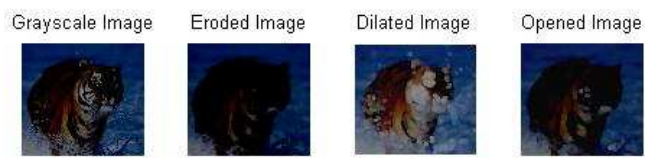

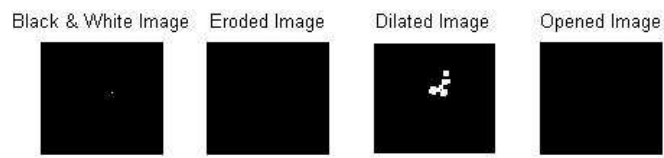

(c8)

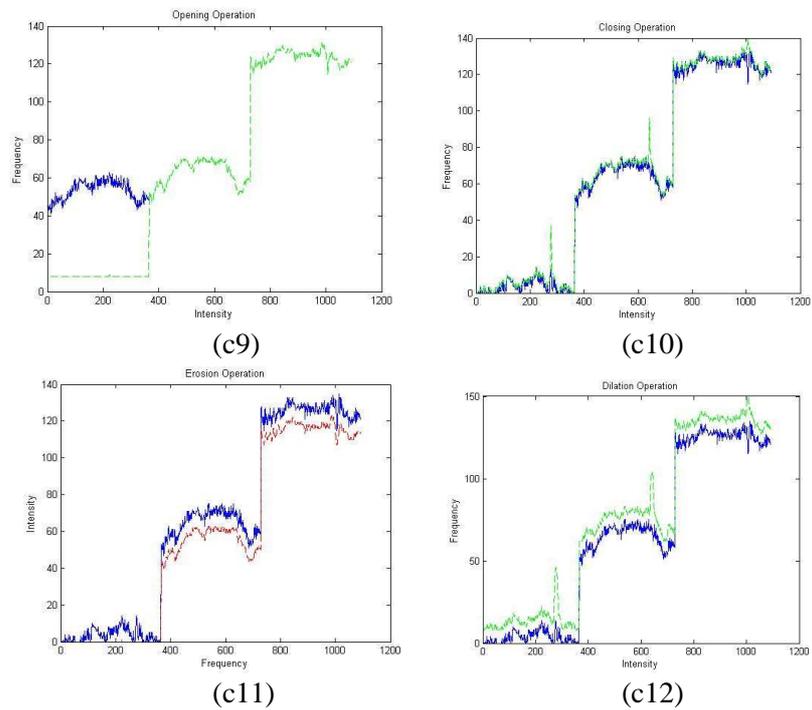

(c9)          (c10)

(c11)          (c12)

Figure 13.Image background detection using block approach and contrast enhancement (c1).Background detection image, (c2). Original image (c3).GRAY image, (c4). IR image (c5).Eroded image, (c6).Dilated image (c7).Block Analysis, (c8).Image Operation (c9).Opening Operation, (c10). Closing Operation, (c11).Erosion Operation, (c11).Dilation Operation

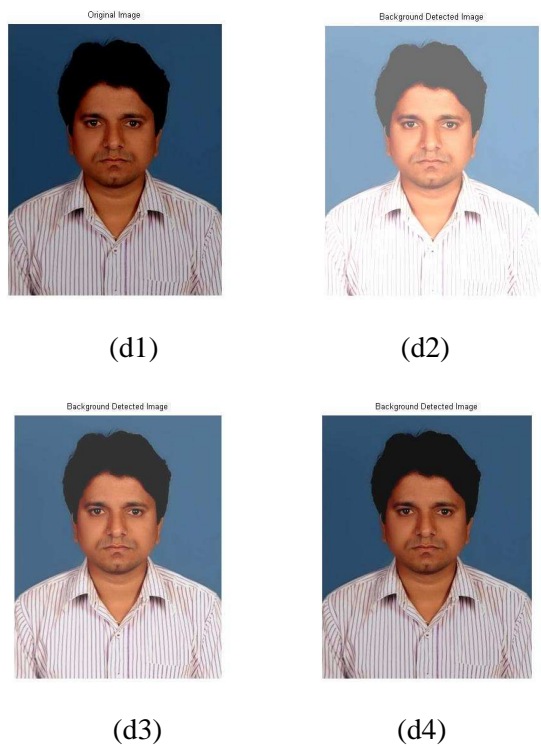

(d1)          (d2)

(d3)          (d4)

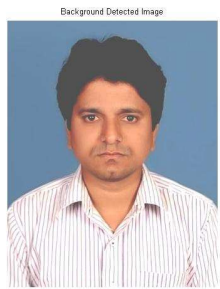 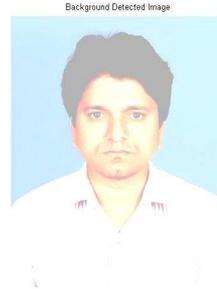

(d5)          (d6)

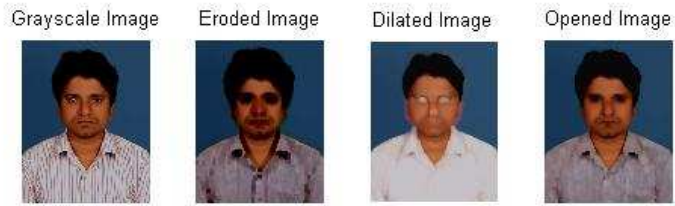

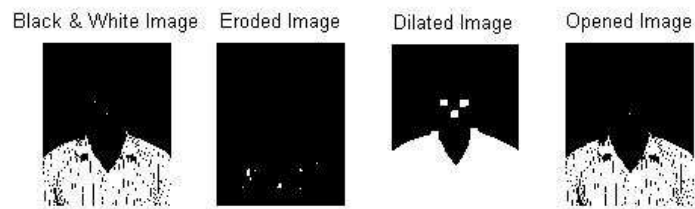

(d7)

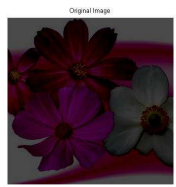 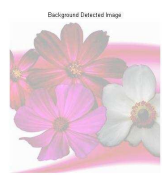

(e1)          (e2)

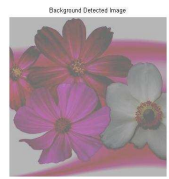 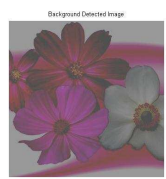

(e3)          (e4)

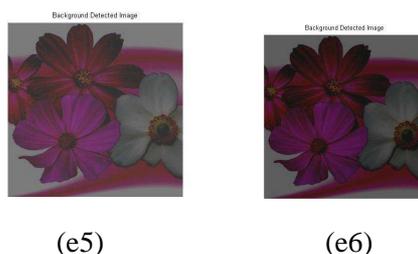

(e5)　　　　　　　(e6)

Figure 14.Image background using the opening by reconstruction with different sizes. (d1) Original image; (d2), (d3), (d4), (d5), (d6) background images and (d7) Image operation obtained after applying equation (22) with structuring element sizes $\mu = 20, 50, 80, 110, and \ 180$ (e1) Original image, (e2), (e3), (e4), (e5), (e6) enhanced images obtained from the application of equation (24).

## 6. CONCLUSION

First, a methodology was introduced to compute an approximation to the background using blocks analysis. This proposal was subsequently extended using mathematical morphology operators. However, a difficulty was detected when the morphological erosion and dilation were employed; therefore, a new proposal to detect the image background was propounded, that is based on the use of morphological connected transformations. Also, morphological contrast enhancement transformations were introduced. Such operators are based on Weber's law notion. The performances of the proposals provided in this work were illustrated by means of several examples throughout the paper. Also, the operators performance employed in this paper were compared with others given in the literature. Finally, a disadvantage of contrast enhancement transformations studied in this paper is that they can only be used satisfactorily in images with poor lighting; in a future work this problem will be considered.

**Authors**

**K.Sreedhar** received the B.Tech. degree in Electronics and Communication Engineering from JNTUH University, Hyderabad, India and M.Tech degree in Communication Systems from JNTUH University, Hyderabad, India . He attended the International Conference on Technology and Innovation at Chennai. He also attended the National Conference at Coimbatore, Tamilnadu, India on INNOVATIVE IN WIRELESS TECHNOLOGY. He is working as a Assistant Professor in Electronics and Communication Engineering department at Vivekananda Institute of Science and Technology, Karimnagar, Andhra Pradesh, India He has a Life Member ship in ISTE. He published four International Research papers. 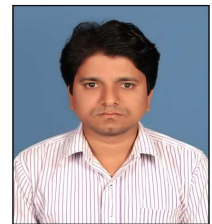

**B.Panlal** received the B.Tech. degree in Electronics and Communication Engineering from JNTUH University, Hyderabad, India and M.Tech degree from KU University, Warangal, India . He has a Life Member ship in ISTE. Presently, He is working at Vaageswari College of engineering, AndhraPradesh, India. 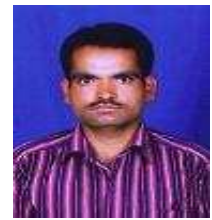